\documentclass[runningheads]{llncs}
\usepackage[T1]{fontenc}
\usepackage{graphicx}
\usepackage{inconsolata}
\usepackage{amsmath,amsfonts}
\usepackage{algorithmic}
\usepackage[section]{algorithm}

\usepackage{graphicx}
\usepackage{enumitem}
\usepackage{array, multirow}
\usepackage{tabularx}
\usepackage{booktabs}
\usepackage{subcaption}

\begin{document}
\title{EvoP: Robust LLM Inference via Evolutionary Pruning}
%
%
%

\author{%
  Shangyu Wu\inst{1} \and
  Hongchao Du\inst{1} \and
  Ying Xiong\inst{2} \thanks{Corresponding author.} \and
  Shuai Chen\inst{3} \and
  Tei-Wei Kuo\inst{4} \and
  Nan Guan\inst{1} \and
  Chun Jason Xue\inst{2}
}
\institute{
City University of Hong Kong \and
Mohamed bin Zayed University of Artificial Intelligence \and
Baidu \and
National Taiwan University 
}

\maketitle              
\begin{abstract}
Large Language Models (LLMs) have achieved remarkable success in natural language processing tasks, but their massive size and computational demands hinder their deployment in resource-constrained environments. 
Existing model pruning methods address this issue by removing redundant structures (e.g., elements, channels, layers) from the model. 
However, these methods employ a heuristic pruning strategy, which leads to suboptimal performance.
Besides, they also ignore the data characteristics when pruning the model.

To overcome these limitations, we propose EvoP, an evolutionary pruning framework for robust LLM inference.
EvoP first presents a cluster-based calibration dataset sampling (CCDS) strategy for creating a more diverse calibration dataset.
EvoP then introduces an evolutionary pruning pattern searching (EPPS) method to find the optimal pruning pattern.
Compared to existing model pruning techniques, EvoP achieves the best performance while maintaining the best efficiency. 
Experiments across different LLMs and different downstream tasks validate the effectiveness of the proposed EvoP, making it a practical and scalable solution for deploying LLMs in real-world applications.

\keywords{Large Language Model \and Layer-wise Model Pruning \and Evolutionary Search.}
\end{abstract}
\section{Introduction}

Large Language Models (LLMs), such as Llama-series~\cite{llama2} and OPT-series~\cite{opt}, have revolutionized natural language processing (NLP) by achieving state-of-the-art performance across a wide range of tasks~\cite{llama,opt,bloom,palm,lamda,glm}.
However, the success of these models comes at a significant computational cost. 
LLM consists of billions of parameters, requiring substantial memory, storage, and energy resources. 
This computational burden limits their deployment in resource-constrained environments. 

To address these challenges, network pruning~\cite{lecun_prun} has emerged as a critical technique for optimizing LLMs by reducing their size while preserving their performance.
Pruning involves selectively removing redundant or less important parameters (e.g., elements~\cite{wanda}, channels~\cite{slicegpt}, or layers~\cite{sleb}) from a pre-trained model, which enables faster inference and lower energy consumption. 
Importantly, pruning is feasible in LLMs because these models are often over-parameterized, containing more parameters than necessary to achieve their performance.


Unfortunately, existing pruning techniques still face different challenges. 
Element-wise pruning achieves the finest-grained pruning to obtain a quite high sparsity, e.g., 50\%, but requires specific computation kernels and hardware support to accelerate the inference.
Channel-wise pruning trades off the hardware efficiency and sparsity. 
However, the complex channel dependency increases the difficulty of determining the pruning channel, thus resulting in performance drops.
Layer-wise pruning adopts heuristic algorithms to search the pruning pattern and maintain the hardware friendliness for faster inference.
But it often fails to find the optimal pruning patterns.

To address the challenges, we follow layer-wise pruning and introduce a novel evolutionary pruning framework, EvoP.
EvoP first presents a cluster-based calibration dataset sampling strategy for collecting a more diverse calibration dataset.
EvoP then combines the diverse calibration dataset with an evolutionary pruning pattern search method to identify the optimal pruning patterns that generalize well across tasks.
Extensive experiments demonstrate that the proposed EvoP outperforms existing state-of-the-art pruning methods of different granularity.
The proposed EvoP also performs well on in-domain and out-of-domain perplexity datasets, showing its better generalization capabilities. 
Codes are available at https://github.com/luffy06/EvoP.

The main contributions of this paper are:
\begin{itemize}
    \item We formulate the network pruning problem and present two key observations, including suboptimal solutions in the algorithm aspect and low data diversity in the dataset aspect.
    \item Based on the observations, we propose the evolutionary pruning framework, EvoP, which includes a cluster-based calibration dataset sampling and an evolutionary pruning pattern search.
    \item Experimental results show that the proposed EvoP can achieve the best performance across five downstream tasks and in-/out-of-domain datasets.
\end{itemize}

\section{Background}
\subsection{Network Pruning}
Network pruning is a key technique in the LLM era, which aims to reduce neural networks' size and computational complexity by removing unnecessary or redundant structures, such as elements~\cite{wanda}, channels~\cite{slicegpt}, or layers~\cite{sleb}. 
Existing methods target different levels of pruning granularity, from fine-grained element-wise pruning to coarse-grained layer-wise removal, offering a trade-off between model compression, hardware efficiency, and task performance. 

This paper formally defines the network pruning problem to understand and unify these approaches better.
Given a sparsity hyperparameter $\theta$, a pre-trained backbone model $\mathcal{M}$ with parameters $W$, and a calibration dataset $\mathcal{D}$ with $n$ data, the goal of network pruning is to find a pruning pattern $p$ in the Pruning Pattern Space $P$ (PPS) that minimizes the loss of the pruned model $\mathcal{M}^\theta$ over the calibration dataset $\mathcal{D}$,
\begin{equation}
    \min_{p\in P}\mathcal{L}(f(\mathcal{M}, p), \mathcal{D}), \text{s.t.} \|p\|_0 \leq \theta \cdot \|W\|_0
\end{equation}
where $\mathcal{L}(\cdot)$ is the loss function, $p\in \{0, 1\}^m$ is a binary vector of zeros and ones, indicating a pruning pattern over $m$ pruning components, which can be weight elements, channels, or layers.
$p[i] = 1$ represents the $i$-th components in the model $\mathcal{M}$ is pruned, and vice versa.
Moreover, $f(, )$ is the function that applies the pruning pattern $p$ to the pre-trained model $\mathcal{M}$ with different pruning techniques.
$\|p\|_0$ and $\|W\|_0$ are the number of non-zeros values in $p$ and the total number of parameters in $W$.

\begin{figure}[t]
\centering
\includegraphics[width=1\linewidth]{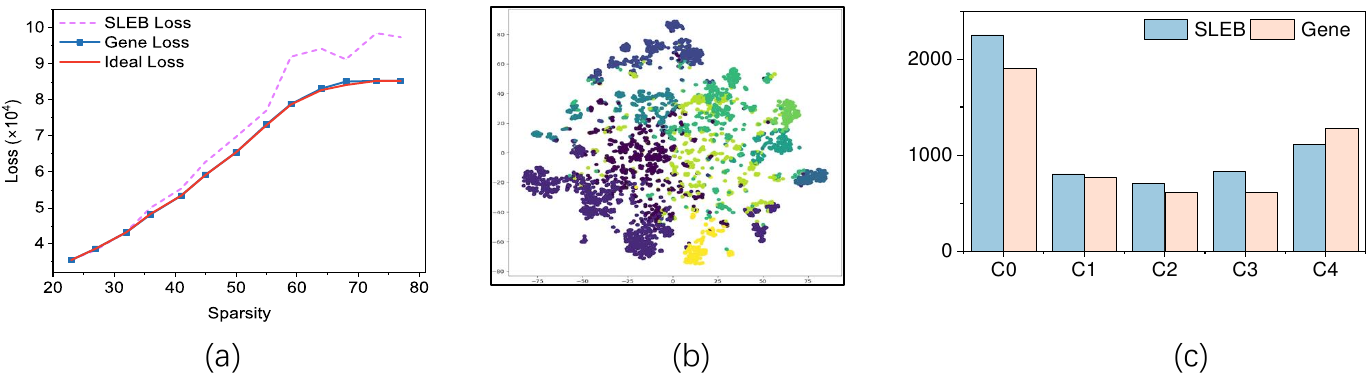}
\caption{Observations. (a)  Perplexity loss of the Tinyllama model on the calibration dataset across different sparsity. The Ideal is the globally optimal solution in each pruning pattern space. The Gene is the solution using the genetic algorithm. SLEB fails to find optimal when the pruning pattern space becomes larger (large sparsity). (b) Visualization of the semantic distribution of the calibration dataset. Each point represents an input sentence. Points with the same color are clustered together by KMeans. (c) Perplexity of SLEB and Gene using different calibration datasets. The sparsity of Tinyllama is set to 50\% for large enough PPS. `C0', `C1', ..., and `C4' are randomly sampled within each cluster.}
\label{fig:1}
\end{figure}

\subsection{Challenges of Different Pruning Techniques}
The above formulation provides a unified framework for understanding and comparing different pruning methods. 
However, each pruning granularity introduces unique challenges.
Element-wise pruning is a kind of unstructured pruning, which is not a hardware-friendly technique, requiring specific computation kernel designs and implementations.
For example, 2:4 pruning~\cite{sparcegpt,wanda,dsnot,nm_sparse1,nm_sparse2} relies on specific sparse matrix multiplication operators and is currently only supported by GPUs with specific architectures, such as NVIDIA Ampere architecture, Hopper.
Besides, the pruning pattern space $P$ of the element-wise pruning can be enormous, resulting in the proposed solutions being generally heuristic algorithms on each weight matrix~\cite{sparsert,neur_prun}.

Channel-wise pruning removes the entire channel for better hardware efficiency~\cite{slicegpt,llmpruner}. 
However, complex dependencies between channels often make it challenging to determine which channel should be pruned, thus leading to substantial performance drops~\cite{att_prun}. 

Recent works~\cite{24arxiv-raee,sleb} pay more attention to coarse-grained pruning, i.e., layer-wise pruning, due to its greater hardware efficiency and better model performance.
Those works can be categorized into two branches: early exit and layer dropping.
The former introduces intermediate exit points within the model and terminates the inference at the predicted exit point~\cite{24arxiv-raee,deebert}, while the latter drops some layers based on the calibration dataset and leaves the pruned model for the inference~\cite{sleb}.
Compared to layer dropping, although early exit seems an easier-to-solve approach to pruning the model, it isn't easy to accelerate the inference in batches and reduce memory overheads in deployment~\cite{sleb}.
Considering the above challenges of different pruning techniques, this paper mainly focuses on optimizing the layer dropping in layer-wise pruning.

\section{Observations}

This section presents two observations regarding algorithms and data, which motivate us to propose corresponding methods.


\begin{figure*}[t]
\centering
\includegraphics[width=0.8\linewidth]{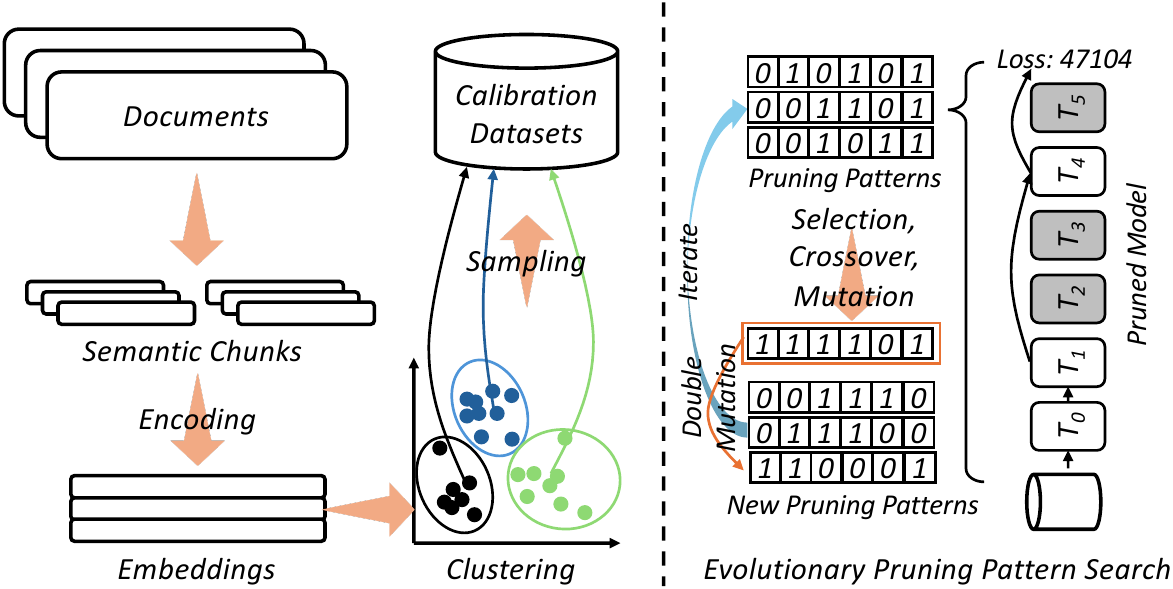}
\caption{The overview of EvoP. The left part shows the cluster-based calibration dataset sampling. The right part presents the evolutionary pruning pattern search.}
\label{fig:overview}
\end{figure*}

\subsection{Observation 1: Sub-optimal Pruning Pattern of Heuristics in Large PPS}

Existing state-of-the-art layer-dropping techniques adopt heuristic algorithms to select layers to drop.
For example, SLEB~\cite{sleb} enumerates all unpruned layers and removes the one where the model without the layer achieves the minimum perplexity loss on the calibration dataset.
Then, SLEB repeats this process until the given sparsity is reached.
This simple yet effective technique is successful because the SLEB algorithms can achieve as optimal as ideal solutions when the pruning pattern space is not large enough (small sparsity).

To better support the above claims, we conducted an analysis experiment on the Tinyllama model~\cite{tinyllama} over the calibration dataset in Figure~\ref{fig:1}(a).
We follow the data selection of SLEB to build the calibration dataset from WikiText-2~\cite{wikitext2}.
We collect three sets of perplexity losses over different sparsities: the SLEB, the Ideal that enumerates the whole pruning pattern space, and the Gene using the genetic algorithm to search.

As expected, when the sparsity is below around 35\%, SLEB can easily find the global optimal 
since the pruning pattern space is small, only seven layers need to be selected at most.
However, when the size of the pruning pattern space increases (larger sparsity), SLEB gradually gets stuck in the local optimal.
Therefore, this motivates us to propose a new algorithm to further optimize the pruning in the scenarios with larger pruning pattern space. 
Noted, the size of the pruning pattern space is not only related to the given sparsity but also depends on the original model size, which can be calculated as the combinatorial number. The number of model layers is the total number of $N$, and the number of pruned layers is the degree $k$, thus the size of the pruning pattern space is $\binom{N}{k}$. 
Therefore, even a smaller sparsity (e.g., 10\%-20\%) can lead to an ample pruning pattern space for larger models.

\subsection{Observation 2: Low Data Diversity of Calibration Dataset}

The above analysis shows the potential to approach the global optimal.
However, the global optimal might differ when pruning on different calibration datasets.
First, we analyze the semantic distribution of WikiText-2 and visualize the distribution in Figure~\ref{fig:1}(b).
We use the BERT-base~\cite{BERT} model to encode each sentence and TSNE to reduce the embedding dimensions.
Figure~\ref{fig:1}(b) demonstrates that some sentences are semantically similar and naturally clustered.
We also use KMeans to cluster those sentences and draw the colors, proving this claim.

Based on the data characteristics, we further show the impact of different calibration datasets on the global optimal pruning patterns.
Since Figure~\ref{fig:1}(a) has demonstrated the Gene can achieve comparable performance to the Ideal, we would regard the Gene's results as the Ideal's.
We collect the perplexity of SLEB and Gene using the samples from each cluster in Figure~\ref{fig:1}(c).
Experimental results show that different calibration datasets may lead to different optimal.
This motivates us to propose a new data sampling method to help the pruning algorithms reach a better optimal.

\section{EvoP: An Evolutionary Pruning Framework}
\label{sec:dyna}

Figure~\ref{fig:overview} shows the overview of EvoP, which consists of two key steps: 
(1) Build a robust calibration dataset based on the Cluster-based Calibration Data Sampling (CCDS), 
and (2) Determine the optimal pruning pattern with the Evolutionary Pruning Pattern Search (EPPS).
Below, we show the details of each step.


\subsection{CCDS: Cluster-based Calibration Dataset Sampling}
To achieve a more robust pruning pattern, EvoP first partition the raw dataset (e.g., WikiText-2~\cite{wikitext2} or C4~\cite{c4}) into $k$ clusters based on input semantic similarity. 
To ensure the clustering effect and semantic integrity, we first divide the dataset into chunks with the same number of sentences and then cluster the data at the chunk level.
BERT~\cite{BERT} is applied to generate the embedding of each chunk since it can get information-rich representations compared to unidirectional pre-trained language models.
Then, clustering algorithms such as KMeans are used to cluster these embeddings.

The generated clusters separate inputs with different computational requirements, allowing for more diverse and representative data sampling.
For each cluster $C_j$ (where j = 1,2,$\dots$,k), We randomly select $n$ representative samples, each of which may contain multiple chunks.
These samples are used to evaluate the performance of different pruning patterns during the search process. 
Sampling across all clusters ensures data diversity, 
improving the robustness of the pruning patterns.
The detailed algorithm is shown in Algorithm~\ref{alg:algo}.

\begin{algorithm}[t]
\caption{Cluster-based Calibration Data Sampling (CCDS)}
\label{alg:algo}
\begin{algorithmic}[1]

\REQUIRE Raw Dataset \( \mathcal{D} \), number of clusters \( k \), number of samples per cluster \( n \), the maximum length of each samples \(len\)
\ENSURE Calibration dataset \( \mathcal{X}\)

\STATE Divide \( \mathcal{D} \) into chunks with the same number of sentences \(\{c_1,c_2,\dots,c_{max}\}\)

\FOR{$i = 1$ \textbf{to} $max$}
    \STATE \(e_i = \text{BERT\_embedding}(c_i) \)
\ENDFOR

\STATE \( \{\mathcal{C}_1, \mathcal{C}_2, \dots, \mathcal{C}_k\} = \text{KMeans}(E,k)\), where \(E=\{e_1,e_2,\dots,e_{max}\}\)


\FOR {each cluster \( \mathcal{C}_j \)} 
     \FOR{$i = 1$ \textbf{to} $n$}
    \WHILE {\(|x_i| < len\)}
        \STATE randomly select a chunk from \( \mathcal{C}_j \) and append to \(x_i\).
    \ENDWHILE
    \STATE Add \(x_i[0:len]\) to \(\mathcal{X}_j\)
    \ENDFOR
\ENDFOR
\RETURN \( \mathcal{X} = \{\mathcal{X}_1,\mathcal{X}_2,\dots,\mathcal{X}_k\}\).

\end{algorithmic}
\end{algorithm}

\begin{algorithm}[t]
\caption{Evolutionary Pruning Pattern Search (EPPS)}
\label{alg:algo2}
\begin{algorithmic}[1]

\REQUIRE Calibration dataset \( \mathcal{X}\), sparsity level \( \theta\), number of generations \( G \), population size \( S \), mutation rate \( \mu \)

\ENSURE Pruning patterns \( \mathbf{p^*} \)
\STATE  Generate \( S \) random pruning patterns \(  \mathcal{P} =\{\mathbf{p}_1, \mathbf{p}_2, \dots, \mathbf{p}_S\} \) that satisfies sparsity \( \theta\), where \( \mathbf{p}_i \in \{0, 1\}^m \).
\FOR {each generation \( g = 1 \) \textbf{to} \( G \)}
    \FOR {each \( \mathbf{p}_i \) in \(\mathcal{P}\)}
        \STATE compute the average loss \( \mathcal{L}(\mathbf{p}_i) \) on \( \mathcal{X}\).
    \ENDFOR
    \STATE  Select the top part of pruning patterns with the lowest loss.
    \STATE Generate new patterns by combining pairs of selected patterns.
    \STATE Randomly flip bits in the new patterns with probability \( \mu \).
    \STATE Adjust the new patterns according to sparsity \(\theta\) and replace \(\mathcal{P}\).
\ENDFOR
\STATE Choose the pruning pattern \( \mathbf{p^*} \) with the lowest loss.
\RETURN The pruning pattern \( \mathbf{p^*}\).

\end{algorithmic}
\end{algorithm}

\subsection{EPPS: Evolutionary Pruning Pattern Search}
Unlike heuristic methods in SLEB~\cite{sleb}, which rely on predefined rules (e.g., magnitude-based pruning), we proposed a revolutionary pruning pattern search algorithm (Algorithm~\ref{alg:algo2}) to explore search space of possible pruning patterns systematically and adaptively. 
First, a population of pruning patterns is initialized, where each pattern is represented as a binary vector $p \in \{0,1\}^{m}$, with 
$m$ is the number of layers in the LLM.
A value of 1 indicates that the layer is pruned, while 0 means it is retained. 
Then, for each pruning pattern $p$, the fitness is computed as the average loss on the $k\times n$ representative samples.
The loss is measured using the task-specific objective (e.g., perplexity loss for language modeling). 
EvoP uses perplexity to measure the loss and select the top patterns with the lowest loss (highest fitness) to generate the next generation. We select the top 30\% in the implementation.
This mimics natural selection, where the fittest individuals survive. 
New pruning patterns are generated by combining (crossover) and randomly altering (mutation) the selected patterns. 
This introduces diversity into the population, enabling the exploration of new pruning configurations. Note that the newly pruning patterns may not meet the sparsity requirements, so we randomly select several corresponding positions to flip until the required sparsity is met (double mutation).
Last, the process repeats for several generations or until convergence. 
The final pruning pattern for the cluster is the one with the lowest loss.

\section{Experiments}
In this section, we first introduce the dataset and the experimental setup.
Then, we present the results of our EvoP and different pruning techniques on five downstream tasks.
We also conduct analysis experiments to show our EvoP and those pruning techniques' in-domain and out-of-domain performance.
Finally, we show the ablation studies.

\begin{table*}[t]
\centering
\begin{tabular}{lccccccccc}
\toprule
\multicolumn{2}{c}{Methods} & Sparsity & SpeedUp & ARC\_E & ARC\_C & HellaSwag & PIQA & Winogrande & Avg \\
\midrule
\multirow{8}{*}{\rotatebox{90}{Llama-2-13b}} 
& Dense & 0\% & 1.00x & 0.82 & 0.53 & 0.61 & 0.80 & 0.75 & 0.70 \\ \cline{2-10}
& Wanda (2:4) & 50\% & 1.00x & 0.71 & 0.35 & 0.43 & 0.72 & 0.62 & 0.57 \\ 
& \multirow{2}{*}{SliceGPT} & 10\% & 0.99x & 0.65 & 0.34 & 0.44 & 0.69 & 0.70 & 0.56 \\
& & 20\% & 1.10x & 0.44 & 0.24 & 0.34 & 0.60 & 0.62 & 0.45 \\
& \multirow{2}{*}{SLEB} & 10\% & 1.10x & \textbf{0.77} & 0.43 & 0.56 & 0.78 & 0.69 & 0.65 \\
& & 20\% & 1.23x & \underline{0.73} & 0.39 & 0.50 & \underline{0.77} & 0.65 & 0.61 \\ \cline{2-10}
& \multirow{2}{*}{EvoP} & 10\% & 1.10x & \textbf{0.77} & \textbf{0.46} & \textbf{0.57} & \textbf{0.79} & \textbf{0.71} & \textbf{0.66} \\
& & 20\% & 1.23x & \underline{0.73} & \underline{0.40} & \underline{0.52} & \underline{0.77} & \underline{0.68} & \underline{0.62} \\ 
\midrule
\multirow{8}{*}{\rotatebox{90}{OPT-13b}}
& Dense & 0\% & 1.00x & 0.71 & 0.35 & 0.52 & 0.76 & 0.67 & 0.60 \\ \cline{2-10}
& Wanda (2:4) & 50\% & 1.00x &  0.63 & 0.28 & 0.42 & 0.71 & 0.60 & 0.53 \\
& \multirow{2}{*}{SliceGPT} & 10\% & 0.93x & 0.67 & 0.32 & 0.49 & 0.74 & 0.65 & 0.57 \\
& & 20\% & 1.01x & 0.60 & 0.31 & 0.44 & 0.69 & 0.63 & 0.53 \\
& \multirow{2}{*}{SLEB} & 10\% & 1.11x & \textbf{0.70} & 0.34 & \textbf{0.52} & \textbf{0.76} & \textbf{0.67} & \textbf{0.60} \\
& & 20\% & 1.24x & 0.65 & 0.30 & 0.47 & 0.74 & 0.64 & 0.56 \\ \cline{2-10}
& \multirow{2}{*}{EvoP} & 10\% & 1.11x & \textbf{0.70} & \textbf{0.35} & \textbf{0.52} & \textbf{0.76} & \textbf{0.67} & \textbf{0.60} \\
& & 20\% & 1.24x & \underline{0.67} & \underline{0.33} & \underline{0.50} & \underline{0.75} & \underline{0.66} & \underline{0.58} \\ 
\midrule
\multirow{8}{*}{\rotatebox{90}{OPT-30b}}
& Dense & 0\% & 1.00x & 0.72 & 0.37 & 0.55 & 0.77 & 0.69 & 0.62 \\ \cline{2-10}
& Wanda (2:4) & 50\% & 1.00x & 0.64 & 0.29 & 0.42 & 0.70 & 0.61 & 0.53 \\
& \multirow{2}{*}{SliceGPT} & 10\% & 0.95x & 0.69 & 0.34 & 0.51 & 0.75 & 0.68 & 0.59 \\
& & 20\% & 1.01x & 0.63 & 0.32 & 0.47 & 0.72 & 0.65 & 0.56 \\
& \multirow{2}{*}{SLEB} & 10\% & 1.12x & 0.71 & 0.36 & 0.53 & \textbf{0.77} & 0.68 & 0.61 \\
& & 20\% & 1.26x & 0.69 & \underline{0.35} & 0.50 & 0.75 & 0.65 & \underline{0.59} \\ \cline{2-10}
& \multirow{2}{*}{EvoP} & 10\% & 1.12x & \textbf{0.72} & \textbf{0.38} & \textbf{0.55} & \textbf{0.77} & \textbf{0.69} & \textbf{0.62} \\
& & 20\% & 1.26x & \underline{0.70} & 0.34 & \underline{0.51} & \underline{0.76} & \underline{0.66} & \underline{0.59} \\
\bottomrule
\end{tabular}
\caption{Performance on five downstream tasks. Results in \textbf{bold} are the best performance with 10\% sparsity, and results in \underline{underline} are the best performance with 20\% sparsity. Since Wanda only supports 50\%, we compare it to models with 10\% and 20\%, respectively. `ARC\_C' and `ARC\_E' represent the tasks ARC\_Easy and ARC\_Challenge.}
\label{tab-down}
\end{table*}

\subsection{Datasets and Experimental Setup}

\textbf{Datasets.} Following SLEB~\cite{sleb}, we also use the library LM-Harness~\cite{eval-harness} to evaluate our EvoP and other pruning techniques on the same five representative datasets.
These datasets cover a wide range of downstream tasks, including reasoning, commonsense understanding, and question answering, making them suitable for evaluating the generalization ability of pruned models.

\noindent\textbf{Models.} We conducted experiments on three large language models, i.e., Llama-2-13b~\cite{llama2}, OPT-13b, and OPT-30b~\cite{opt}.
These models can serve as representative benchmarks to analyze the effectiveness of our EvoP across different architectures and scales.

\noindent\textbf{Baselines.} We choose the state-of-the-art techniques of each kind of network pruning,
\begin{itemize}[nosep]
    \item \textbf{Wanda~\cite{wanda}:} A superior \textit{element-wise pruning} that prunes weights with the smallest magnitudes multiplied by the corresponding input activations.
    \item \textbf{SliceGPT~\cite{slicegpt}:} An advanced \textit{channel-wise pruning} that prunes the columns or rows of weights.
    \item \textbf{SLEB~\cite{sleb}:} A surpassing \textit{layer-wise pruning} that removes some layers or blocks of the model.
\end{itemize}
Noted, all baselines, including our EvoP, require no retraining or weight update, and the pruned LLM can be used as is.

\noindent\textbf{Implementation Details.} The proposed EvoP was implemented using the PyTorch and Transformer frameworks.
We evaluated the baseline with Llama-2-13b and OPT-13b on one NVIDIA A100 GPU with 40GB memory and 
the baseline with OPT-30b on one NVIDIA A800 GPU with 80GB memory.
The size of the calibration dataset includes 5 samples with 2048 tokens each, which is equivalent to the dataset used in SLEB~\cite{sleb}.
The number of clusters in EvoP is set to 5, and the max number of iterations of EPPS is set to 100 for models with 13 billion parameters and 200 for models with 30 billion parameters.

\begin{table*}[t]
\centering
\begin{tabular}{lccccccc}
\toprule
\multirow{2}{*}{Methods} & \multirow{2}{*}{Sparsity} & \multicolumn{3}{c}{WikiText2-WikiText2} & \multicolumn{3}{c}{C4-C4} \\
& & Llama-2-13b & OPT-13b & OPT-30b & Llama-2-13b & OPT-13b & OPT-30b \\
\midrule
Dense & 0\% & 4.57 & 10.13 & 9.56 & 6.52 & 12.07 & 11.46 \\ \hline
Wanda & 50\% & 7.93 & 15.17 & 14.93 & 12.28 & 16.97 & 17.50 \\ 
SliceGPT & 10\% & 6.57 & 11.08 & 10.51 & 10.49 & 13.73 & 12.73 \\
SliceGPT & 20\% & 10.26 & 13.19 & 12.09 & 16.93 & 17.50 & 15.88 \\ 
SLEB & 10\% & 5.37 & 10.30 & 10.19 & 7.57 & 12.45 & 11.57 \\
SLEB & 20\% & 6.68 & 12.82 & 11.07 &  \underline{9.00} & 13.97 & 12.17\\
\midrule
EvoP & 10\% & \textbf{5.24} &  \textbf{10.16} &  \textbf{9.75} &  \textbf{7.51} &  \textbf{12.35} &  \textbf{11.55} \\
EvoP & 20\% &  \underline{6.30} &  \underline{11.56} &  \underline{11.01} & 9.01 &  \underline{13.63} &  \underline{12.10} \\
\bottomrule
\end{tabular}
\caption{In-domain perplexity results. Results in \textbf{bold} are the best performance with 10\% sparsity, and results in \underline{underline} are the best performance with 20\% sparsity.}
\label{tab-id-ppl}
\end{table*}

\begin{table*}[t]
\centering
\begin{tabular}{lccccccc}
\toprule
\multirow{2}{*}{Methods} & \multirow{2}{*}{Sparsity} & \multicolumn{3}{c}{WikiText2-C4} & \multicolumn{3}{c}{C4-WikiText2} \\
& & Llama-2-13b & OPT-13b & OPT-30b & Llama-2-13b & OPT-13b & OPT-30b \\
\midrule
Dense & 0\% & 6.52 & 12.07 & 11.46 & 4.57 & 10.13 & 9.56 \\
\midrule
Wanda & 50\% & 12.53 & 18.94 & 22.08 & 8.58 & 17.51 & 16.42 \\ 
SliceGPT & 10\% & 18.42 & 14.04 & 13.09 & 9.69 & 11.91 & 11.26 \\
SliceGPT & 20\% & 43.77 & 19.24 & 16.74 & 22.72 & 17.16 & 15.03 \\ 
SLEB & 10\% & 7.62 & 12.44 & 12.36 & 5.33 & 10.56 & \textbf{9.65} \\
SLEB & 20\% & 9.38 & 15.22 & 13.76 & \underline{6.31} & 12.40 & 10.22 \\
\midrule
EvoP & 10\% & \textbf{7.55} & \textbf{12.37} & \textbf{11.68} & \textbf{5.26} & \textbf{10.38} & \textbf{9.65} \\
EvoP & 20\% & \underline{8.97} & \underline{13.69} & \underline{13.14} & 6.33 & \underline{11.42} & \underline{10.15} \\
\bottomrule
\end{tabular}
\caption{Out-of-domain perplexity results. Results in \textbf{bold} are the best performance with 10\% sparsity, and results in \underline{underline} are the best performance with 20\% sparsity.}
\label{tab-ood-ppl}
\end{table*}

\subsection{Results on Downstream Tasks}

Table~\ref{tab-down} shows the performance on six downstream tasks.
Overall, experimental results demonstrate that the proposed EvoP outperforms existing state-of-the-art pruning techniques on the representative downstream tasks across different sparsity levels.
Compared to the lower sparsity (10\%), the higher sparsity (20\%) represents a larger pruning pattern space, which increases the difficulty of the network pruning problem. 
From the results, our EvoP achieves a better performance on higher-sparsity cases, consistently achieving better performance compared to baselines.
The above results indicate EvoP's robustness in balancing sparsity and performance.

Table~\ref{tab-down} also presents the inference speedup of the pruned model with different pruning techniques compared to the dense model.
Wanda achieves the same inference speed as that of the dense model.
This is due to the fact that Wanda requires customized implementations using hardware-friendly codes.
Otherwise, running Wanda's pruned model is like running the dense model with half-zero weights.
SliceGPT with lower sparsity runs a bit slower than the dense model but a bit faster when with high sparsity.
This might lie in that the extra overheads of SliceGPT are quite large; only when applying large sparsity would the efficiency benefits be dominant.
For SLEB and our EvoP, these two are all layer-wise pruning, which only removes some layers.
The pruned model would be quite hardware-friendly; thus, the speedup gains as much as the given sparsity.
Our EvoP only removes layers according to the searched pruning pattern. 
Thus, its speedup is exactly the same as that of SLEB.

\subsection{Results on Language Modeling}
Table~\ref{tab-id-ppl} and Table~\ref{tab-ood-ppl} show the in-domain and out-of-domain perplexity results on the wikitext-2 and C4 datasets.
Experiments in Table~\ref{tab-id-ppl} show that our EvoP consistently outperforms other pruning techniques, especially on the sparsity of 10\%, demonstrating its effectiveness in maintaining in-domain performance.
Similar conclusions can also be drawn from out-of-domain results. 

\subsection{Ablation Study}


\begin{table}[t]
\centering
\begin{tabular}{lccc}
\toprule
Methods & Sparsity & L-13 & O-13 \\
\midrule
EvoP & 10\% & 7.51 & 12.35 \\
\quad w/o CCDS & 10\% & 7.62 & 12.44  \\
\quad w/o EPPS & 10\% & 7.55 & 12.37   \\
EvoP & 20\% & 9.01 & 13.63 \\
\quad w/o CCDS & 20\% & 9.21 & 13.76 \\
\quad w/o EPPS & 20\% & 9.31 & 13.78    \\
\bottomrule
\end{tabular}
\caption{Perplexity results on C4 datasets using the calibration dataset sampled from WikiText-2. The heuristic algorithm used in SLEB is adopted for EvoP w/o EPPS. `L-13' and `O-13' represents Llama-2-13b model and OPT-13b model.}
\label{tab-ablation}
\end{table}

We conducted the ablation study of different components with different sparsity on the C4 calibration dataset.
As shown in Table~\ref{tab-ablation}, CCDS can help find a better pruning pattern on the low sparsity level, while EPPS contributes more to the performance improvements on the high sparsity level.
This may be due to the fact that at the low sparsity level, the pruning pattern space is not large enough, and heuristic algorithms can easily find the same optimal pruning pattern as the EPPS.
However, at the high sparsity level, a large pruning pattern space is better for demonstrating the effectiveness of the EPPS, thus achieving more performance improvements.


\section{Related Work}

Network pruning, also known as sparsity techniques, refers to methods employed to compress language models by eliminating less significant structures within the network, such as individual weights, neurons, or layers~\cite{sparcegpt,slicegpt,sleb}. The primary objectives are to reduce the model’s size, enhance inference speed, and decrease memory consumption while preserving or only marginally affecting its performance. Recent studies have investigated pruning at various granularities, ranging from coarse to fine. At the coarse-grained level, pruning methods include layer-wise~\cite{sleb}, attention heads~\cite{att_prun}, channel-wise pruning~\cite{slicegpt,llmpruner}, and neurons~\cite{neur_prun}. At the fine-grained level, techniques such as N:M sparsity~\cite{sparcegpt,wanda,dsnot,nm_sparse1,nm_sparse2} and individual weight pruning~\cite{sparsert} have been explored.
This paper mainly explored but was not limited to layer-wise network pruning.
\section{Conclusion}
In this paper, we proposed EvoP, an evolutionary pruning framework for robust LLM inference.
EvoP leverages a cluster-based calibration dataset sampling and an evolutionary pruning pattern searching to jointly help further search the optimal pruning pattern for the model.
Comprehensive experiments demonstrate the best performance of EvoP and high inference efficiency.

\bibliographystyle{splncs04}
\bibliography{ref}

\begin{thebibliography}{10}
\providecommand{\url}[1]{\texttt{#1}}
\providecommand{\urlprefix}{URL }
\providecommand{\doi}[1]{https://doi.org/#1}

\bibitem{slicegpt}
Ashkboos, S., et~al.: Slicegpt: Compress large language models by deleting rows and columns. In: Proceedings of {ICLR} (2024)

\bibitem{palm}
Chowdhery, A., et~al.: Palm: Scaling language modeling with pathways. Journal of Machine Learning Research  \textbf{24},  240:1--240:113 (2023)

\bibitem{BERT}
Devlin, J., et~al.: {BERT:} pre-training of deep bidirectional transformers for language understanding. In: Proceedings of {NAACL-HLT}. pp. 4171--4186 (2019)

\bibitem{sparcegpt}
Frantar, E., Alistarh, D.: Sparsegpt: Massive language models can be accurately pruned in one-shot. In: Proceedings of {ICML} (2023)

\bibitem{eval-harness}
Gao, L., et~al.: A framework for few-shot language model evaluation (07 2024)

\bibitem{24arxiv-raee}
Huang, L., et~al.: {RAEE:} {A} training-free retrieval-augmented early exiting framework for efficient inference. CoRR  (2024)

\bibitem{neur_prun}
Jiang, C., et~al.: Efficient {DNN} neuron pruning by minimizing layer-wise nonlinear reconstruction error. In: Proceedings of {IJCAI}. pp. 2298--2304 (2018)

\bibitem{lecun_prun}
LeCun, Y., Denker, J.S., Solla, S.A.: Optimal brain damage. In: Proceedings of {NeurIPS}. pp. 598--605 (1989)

\bibitem{llmpruner}
Ma, X., Fang, G., Wang, X.: Llm-pruner: On the structural pruning of large language models. In: Proceedings of NeurIPS (2023)

\bibitem{wikitext2}
Merity, S., et~al.: Pointer sentinel mixture models. In: Proceedings of {ICLR} (2017)

\bibitem{c4}
Raffel, C., et~al.: Exploring the limits of transfer learning with a unified text-to-text transformer. Journal of Machine Learning Research  \textbf{21},  140:1--140:67 (2020)

\bibitem{bloom}
Scao, T.L., et~al.: {BLOOM:} {A} 176b-parameter open-access multilingual language model. CoRR  (2022)

\bibitem{sleb}
Song, J., et~al.: {SLEB:} streamlining llms through redundancy verification and elimination of transformer blocks. In: Proceedings of {ICML} (2024)

\bibitem{wanda}
Sun, M., et~al.: A simple and effective pruning approach for large language models. In: Proceedings of {ICLR} (2024)

\bibitem{lamda}
Thoppilan, R., et~al.: Lamda: Language models for dialog applications. CoRR  (2022)

\bibitem{llama2}
Touvron, H., et~al.: Llama 2: Open foundation and fine-tuned chat models. CoRR  (2023)

\bibitem{llama}
Touvron, H., et~al.: Llama: Open and efficient foundation language models. CoRR  (2023)

\bibitem{att_prun}
Voita, E., et~al.: Analyzing multi-head self-attention: Specialized heads do the heavy lifting, the rest can be pruned. In: Proceedings of {ACL}. pp. 5797--5808 (2019)

\bibitem{sparsert}
Wang, Z.: Sparsert: Accelerating unstructured sparsity on gpus for deep learning inference. In: Proceedings of {PACT}. pp. 31--42 (2020)

\bibitem{deebert}
Xin, J., et~al.: Deebert: Dynamic early exiting for accelerating {BERT} inference. In: Proceedings of {ACL}. pp. 2246--2251 (2020)

\bibitem{glm}
Zeng, A., et~al.: {GLM-130B:} an open bilingual pre-trained model. In: Proceedings of {ICLR} (2023)

\bibitem{tinyllama}
Zhang, P., et~al.: Tinyllama: An open-source small language model. CoRR  (2024)

\bibitem{opt}
Zhang, S., et~al.: {OPT:} open pre-trained transformer language models. CoRR  (2022)

\bibitem{nm_sparse2}
Zhang, Y., et~al.: Learning best combination for efficient {N:} {M} sparsity. In: Proceedings of NeurIPS (2022)

\bibitem{dsnot}
Zhang, Y., et~al.: Dynamic sparse no training: Training-free fine-tuning for sparse llms. In: Proceedings of {ICLR} (2024)

\bibitem{nm_sparse1}
Zhou, A., et~al.: Learning {N:} {M} fine-grained structured sparse neural networks from scratch. In: Proceedings of {ICLR} (2021)

\end{thebibliography}
\end{document}